\documentclass[conference]{IEEEtran}
\IEEEoverridecommandlockouts

\usepackage{subcaption}
\usepackage{epsfig}
\usepackage{adjustbox}
\usepackage{multirow}
\usepackage{booktabs}

\usepackage{cite}
\usepackage{amsmath,amssymb,amsfonts}
\usepackage{algorithmic}
\usepackage{graphicx}
\usepackage{textcomp}
\usepackage{xcolor}
\usepackage{hyperref}
\def\BibTeX{{\rm B\kern-.05em{\sc i\kern-.025em b}\kern-.08em
    T\kern-.1667em\lower.7ex\hbox{E}\kern-.125emX}}
\begin{document}

\title{Benchmarking of Cancelable Biometrics \\for Deep Templates
}

\author{
Hatef Otroshi Shahreza$^{1,2}$, Pietro Melzi$^{3}$,  Dailé Osorio-Roig$^{4}$, Christian Rathgeb$^{4}$, \\\vspace{5pt} Christoph Busch$^{4,5}$, S\'{e}bastien Marcel$^{1,6}$, Ruben Tolosana$^{3}$, and Ruben Vera-Rodriguez$^{3}$ \\
$^{1}$Biometrics Security and Privacy Group, Idiap Research Institute, Switzerland\\
$^{2}$School of Engineering, \'{E}cole Polytechnique F\'{e}d\'{e}rale de Lausanne (EPFL), Switzerland\\
$^{3}$Biometrics and Data Pattern Analytics (BiDA) Lab, Universidad Autonoma de Madrid (UAM), Spain\\
$^{4}$Biometrics and Internet Security Research Group (da/sec), Hochschule Darmstadt (HDA), Germany\\
$^{5}$Norwegian Biometrics Laboratory (NBL), Norwegian University of Science and Technology (NTNU), Norway\\
$^{6}$School of Criminal Justice, Universit\'{e} de Lausanne (UNIL), Switzerland\\
}

\maketitle

\begin{abstract}
In this paper, we benchmark several cancelable biometrics (CB) schemes on different biometric characteristics. We consider BioHashing, Multi-Layer Perceptron (MLP) Hashing, Bloom Filters, and two schemes based on Index-of-Maximum (IoM) Hashing (i.e., IoM-URP and IoM-GRP). In addition to the mentioned CB schemes, we introduce a CB scheme (as a baseline) based on user-specific random transformations followed by binarization. We evaluate the unlinkability, irreversibility, and recognition performance (which are the required criteria by the ISO/IEC 24745 standard) of these CB schemes on deep learning based templates extracted from different physiological and behavioral biometric characteristics including face, voice, finger vein, and iris. In addition, we provide an open-source implementation of all the experiments presented to facilitate the reproducibility of our results.
\end{abstract}


\section{Introduction}
The templates extracted from  biometric data (e.g., face, voice, finger vein, etc.) in biometric recognition systems generally include privacy-sensitive information about the identities of individuals enrolled in the system. 
Data protection frameworks, such as the EU General Data Protection Regulation (GDPR) \cite{GDPR}, also classify biometric data as sensitive information which requires protection.
In addition, it has been shown that an adversary can reconstruct approximations of the underlying face images from their facial templates \cite{mai2018reconstruction,template_inversion_icip2022}. Similarly, other biometric characteristics can also be reconstructed from their corresponding templates, e.g., vascular images  \cite{kauba2021inverse} or fingerprint images \cite{cappelli2007fingerprint}.

To protect biometric templates, several biometric template protection (BTP) schemes have been proposed \cite{rathgeb2011survey,nandakumar2015biometric,patel2015cancelable}, commonly categorized as \textit{cancelable biometrics} (CB) and \textit{biometric cryptosystems}. In CB schemes, a transformation function (which is dependent of a key) is used to generate protected templates, and then recognition is based on the comparison of protected templates. In biometric cryptosystems, a key is either bound with or generated from a biometric template, and then recognition is based on the correct retrieval or generation of the key \cite{rathgeb2022deep,uludag2004biometric}. In general, there are  four main requirements defined by the ISO/IEC 24745 standard \cite{ISO24745} for each BTP scheme:
\begin{itemize}
	\item \textit{Renewability}: If the protected template of a subject is leaked, it should be possible to revoke existing protected template and generate a new protected template.
	\item \textit{Irreversibility}: If a protected biometric template is compromised, it should be computationally infeasible to reconstruct the original (unprotected) biometric template.
	\item \textit{Unlinkability}: 
	It should not be feasible to determine if two or more protected templates stem from the same (unprotected) biometric template.
	\item \textit{Performance preservation}: The template protection scheme should not significantly degrade the biometric recognition performance.
\end{itemize}

\begin{figure}[!t]
    \centering
\includegraphics[width=\linewidth]{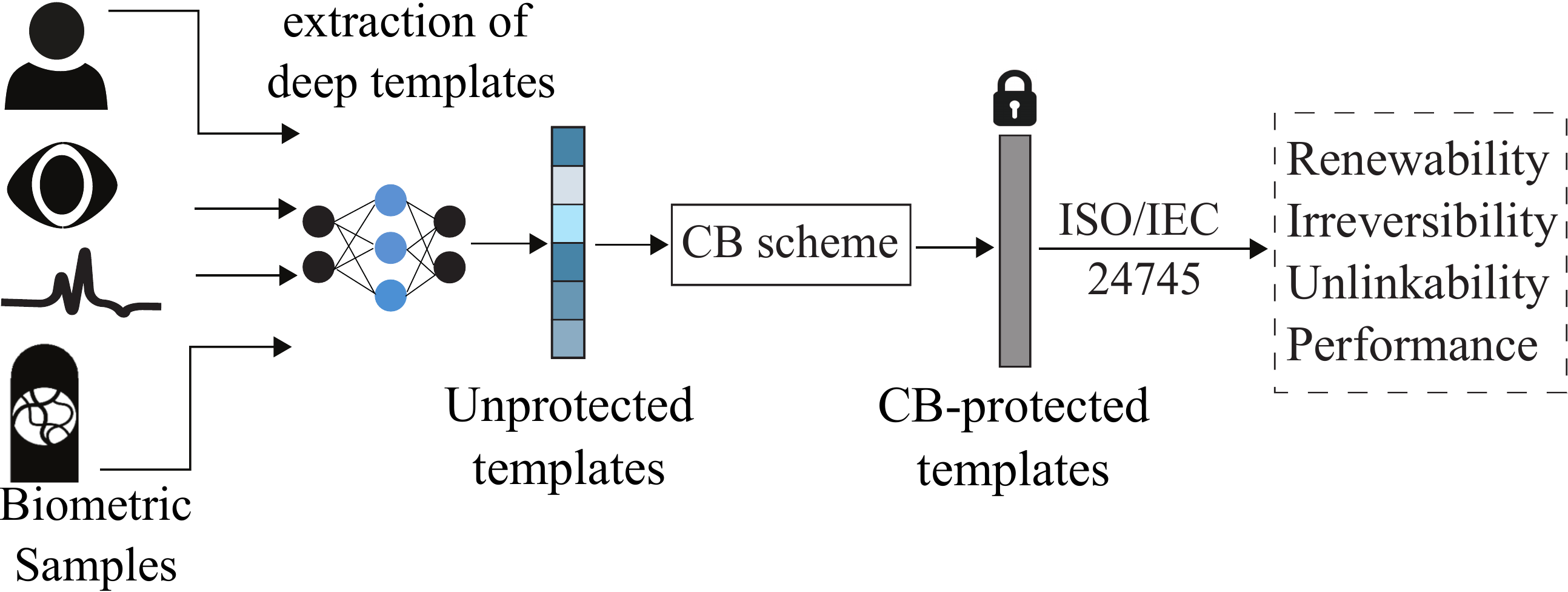}
    \caption{Conceptual overview of benchmarking cancelable biometrics for deep templates.
    }   
    \label{fig:overview-benchmark}
    \vspace{-12pt}
\end{figure}

In this paper, we focus on CB schemes and benchmark several existing methods based on the aforementioned requirements defined in the ISO/IEC 24745 standard on biometric information protection. We consider BioHashing \cite{jin2004biohashing}, Multi-Layer Perceptron (MLP) Hashing \cite{MLPHash}, Bloom Filters \cite{rathgeb2013alignment}, and two schemes based on Index-of-Maximum (IoM) Hashing \cite{jin2017ranking}   (i.e., uniformly random permutation-based hashing,  shortly IoM-URP, and Gaussian random projection-based hashing, shortly IoM-GRP). 
In addition to the mentioned CB schemes, we introduce a CB scheme (as a baseline), dubbed Rand-Hash, based on user-specific random transformation, including random permutation, random scale, and random sign flip, followed by binarization. 
In our experiments, we consider different physiological and behavioral biometric characteristics including face, voice, finger vein, and iris. For each biometric characteristic, we use state-of-the-art (SOTA) feature extraction models in the field which are based on deep neural networks (DNNs).
In a nutshell, the main contribution of this paper is to comprehensively benchmark several CB schemes on DNN-based templates extracted using SOTA models from different biometric characteristics by evaluating unlinkability, irreversibility, and recognition performance. To the best of the authors' knowledge, this work represents the first comprehensive benchmark of CB schemes for DNN-based templates across different biometric characteristics.
The source code of all our experiments are publicly available, and therefore all the results are fully reproducible.

The remainder of this paper is structured as follows. In Section \ref{sec:benchmarking}, we describe our benchmarking framework including the datasets and models for biometric characteristics and also the evaluation metrics. In Section \ref{sec:experiments}, we report our experimental results and benchmark different CB schemes. Finally, the paper is concluded in Section \ref{sec:conclusion}.

\section{Benchmarking Framework}\label{sec:benchmarking}
Fig.~\ref{fig:overview-benchmark} shows a conceptual overview to evaluate different CB schemes applied on deep templates. In this context, different DNN-based feature extractors (Table \ref{tab:modalities}) representing the current state-of-the-art for biometric recognition have been employed. Popular biometric characteristics (Section \ref{subsec:benchmarking:modalities}) have been considered in this analysis. Initially, deep templates are extracted from biometric samples. Then, protected templates are generated using different CB schemes. Finally, different requirements\footnote{Renewability is inherently satisfied due to the application of the key in CB schemes. Therefore, we do not evaluate the renewability of CB schemes.} (i.e., recognition performance, unlinkability, and irreversibility)  defined by the ISO/IEC 24745 standard \cite{ISO24745} are analysed and evaluated in the benchmark (Section \ref{subsec:benchmarking:metrics}).


\begin{table*}[!t]
\begin{center}
    \caption{Summary of feature extraction models, datasets, numbers of mated and non-mated comparisons, and biometric performance in terms of Equal Error Rate (EER) achieved for different biometric characteristics in biometric verification.\vspace{-3pt}}\label{tab:modalities}
        \begin{tabular}{lcccccc}
        \toprule
        \textbf{Characteristic} & \textbf{Model} & \textbf{Dataset} & \textbf{\# Subjects} & \textbf{\# Mated} & \textbf{\# Non-mated} & \textbf{EER} \\
        \midrule
        Face & ArcFace & MOBIO (face)  & 150        & 1,516,300 & 22,952 & 0.02\%\\
        Voice & ECAPA-TDNN & MOBIO (voice) & 150        & 1,516,300 & 22,952 & 6.64\%\\
        Finger Vein & Modified Densenet-161 & SDUMLA &318   &9,540 & 100,806 & 0.32\%\\
        Iris  & Modified Densenet-201 & CASIA Thousand &  457  &13,710	& 207,956 & 2.05\% \\
        \bottomrule
        \vspace{-16pt}
        \end{tabular}
    \end{center}
    \vspace{-2ex}
\end{table*}

\begin{figure}[t]
	\centering
	\begin{subfigure}{\columnwidth}
		\includegraphics[width=0.3\columnwidth]{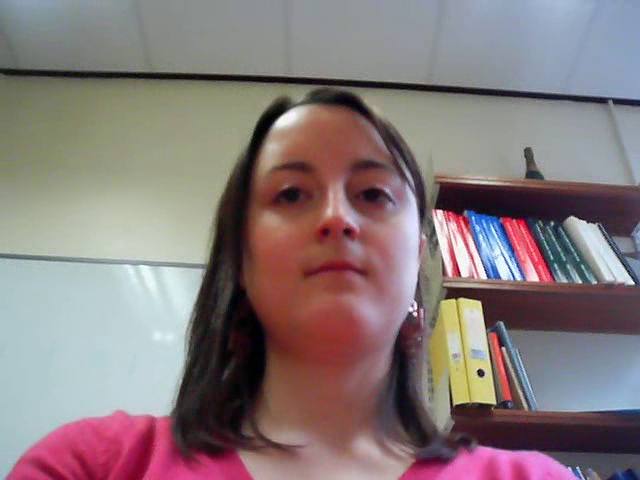}\hfil
		\includegraphics[width=0.3\columnwidth]{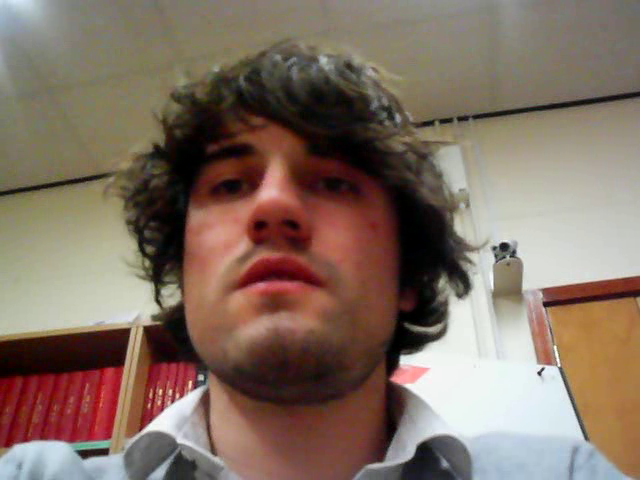}\hfil 
		\includegraphics[width=0.3\columnwidth]{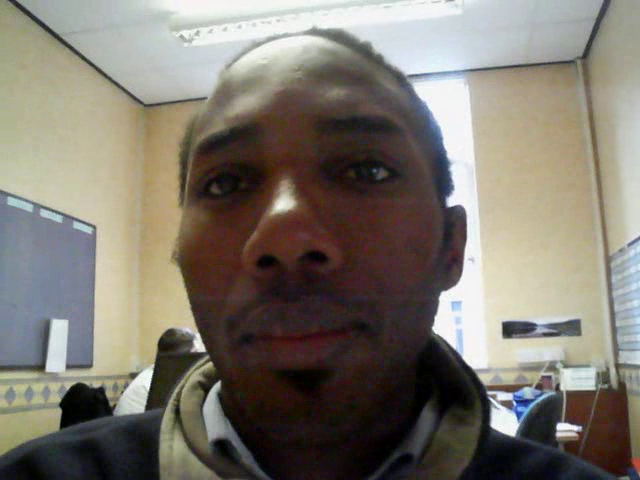}
		\caption{MOBIO (face)}
	\end{subfigure}\\
	
	\begin{subfigure}{\columnwidth}
		\includegraphics[width=0.3\columnwidth]{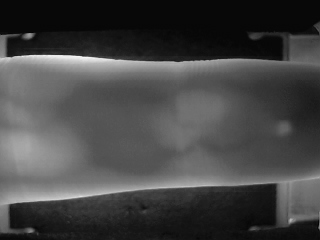}\hfil 
		\includegraphics[width=0.3\columnwidth]{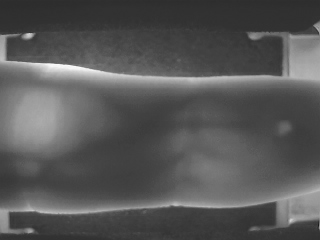}\hfil 
		\includegraphics[width=0.3\columnwidth]{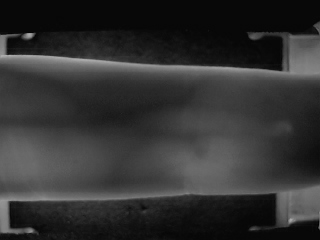}
		\caption{SDUMLA (finger-vein)}
	\end{subfigure}\\
	
	\begin{subfigure}{\columnwidth}
		\includegraphics[width=0.3\columnwidth]{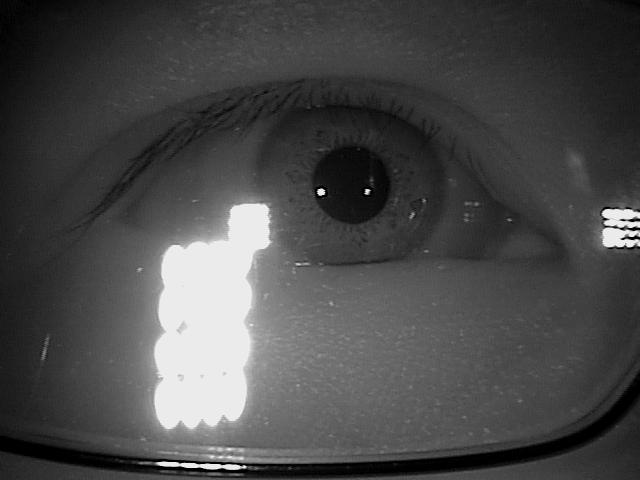}\hfil 
		\includegraphics[width=0.3\columnwidth]{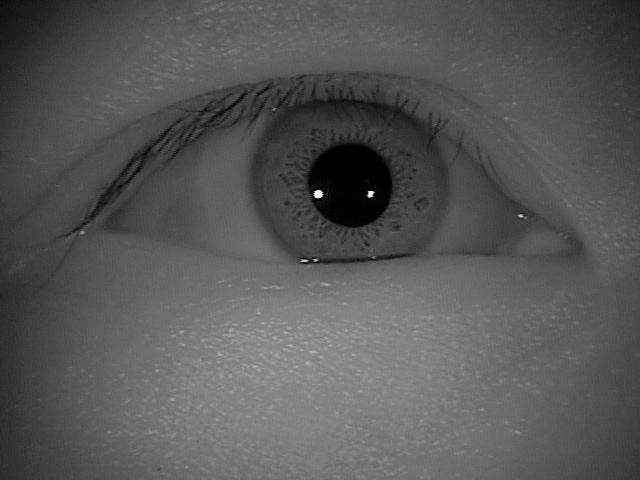}\hfil
		\includegraphics[width=0.3\columnwidth]{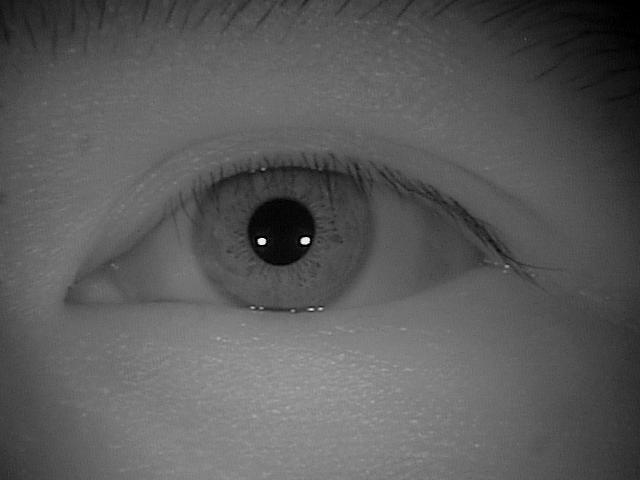}
		\caption{CASIA Thousand (iris)}
	\end{subfigure}\\
	\caption{Sample images of datasets from different biometric characteristics.}
	\label{fig:sample-databases}
	\vspace{-15pt}
\end{figure}

\subsection{Biometric Characteristics}\label{subsec:benchmarking:modalities}
We use different biometric characteristics and implement separate biometric pipelines for each characteristic. Table \ref{tab:modalities} summarises the feature extraction model, the dataset used for each biometric characteristic, the number of mated and not-mated comparisons, and the system verification performance achieved in terms of Equal Error Rate (EER). Figure~\ref{fig:sample-databases} also shows sample images from different datasets.

\subsubsection{Face}
For face recognition, we use the MOBIO \cite{MOBIO} dataset, which is a bimodal dataset including face and voice data acquired using  mobile devices from 150 individuals in 12 sessions. In each session, 6-11 face/voice samples are captured from each individual. To extract deep features from the MOBIO (face)  dataset, we use the ArcFace \cite{deng2018arcface} model.

\subsubsection{Voice}
For voice (speaker) recognition, we use the voice data of the MOBIO dataset (the same dataset described for face), and extract deep features using the ECAPA-TDNN \cite{desplanques2020ecapa} model.

\subsubsection{Finger Vein}
For finger vein recognition, we use the SDUMLA~\cite{Yin-sdumla-2011} dataset which consists of finger vein images of 106 individuals. 
For each individual, 6 instances (index, middle and ring fingers of both hands) are considered, and for each instance 6 samples are captured. We assume each instance as a different subject.
For feature extraction, we use the modified version of DenseNet-161 based on the approach proposed in~\cite{Kuzu-modified-densenet161-2021}. We trained this model on training set (including 53 individuals) and applied on the testing set (including the remaining 53 individuals) according to the protocol in~\cite{Kuzu-modified-densenet161-2021}).  

\subsubsection{Iris} \label{sub:iris}
We consider the CASIA Thousand database \cite{casia}, composed of 1000 individuals, each one represented with 10 images from both their right and left eyes. To extract deep features from iris images, we use the DenseNet-201 network proposed in \cite{hafner_deep_2021}, specifically fine-tuned for iris recognition with the samples of the first 750 individuals of CASIA Thousand. We pre-process the samples of the remaining 250 individuals according to the procedure proposed in \cite{hafner_deep_2021}. Compared to the other biometric characteristics, iris modality requires additional steps for data selection. Samples that contain glasses are identified according to the method proposed in \cite{8411223} and filtered out. After a manual check of pre-processed images, we also discard three samples that provide an incorrect segmentation. Then, we consider as belonging to different subjects the samples obtained from the right and left eyes of the same original individuals. 
We rule out subjects with less than six samples, and limit to the first six the set of samples considered for the remaining subjects.

	

\subsection{Evaluation metrics}\label{subsec:benchmarking:metrics}
In this section, we describe the metrics used to evaluate the unlinkability, irreversibility, and recognition performance in our benchmark. 
We apply the same metrics to all CB schemes, which allows for a direct comparison across different characteristics.
To obtain a fair comparison of CB schemes, as far as possible, we generate protected templates of the same length across different CB schemes in each characteristic. 

\subsubsection{Recognition Performance}
To evaluate the recognition performance of protected templates, we only consider verification and calculate the Equal Error Rate (EER) as well as the False Non-Match Rate (FNMR) at the decision thresholds corresponding to False Match Rates (FMRs) of $1\%$ and $0.1\%$. We also plot the Detection Error Tradeoff (DET) curves. We evaluate the recognition performance in two scenarios: 
\begin{itemize}
    \item \textit{normal:} it is the expected case in practice, we generate protected templates with user-specific keys.
    \item \textit{stolen-token} (shortly \textit{stolen}): we assume that keys are disclosed, hence we evaluate the recognition performance considering the same key for each user.
\end{itemize}
To evaluate the recognition performance, we consider all possible combinations of samples for mated comparisons. For non-mated comparisons, we consider all possible pairs of subjects and use the first sample for each subject in the dataset. In case of iris, mated and non-mated comparisons are generated after performing data selection as described in Section \ref{sub:iris}. We remember that right and left eyes of the same individual are considered as different subjects in the experiments, but no non-mated comparisons are generated from them.

\subsubsection{Unlinkability}
To evaluate unlinkability of CB schemes, we first generate mated and non-mated template pairs with sample-specific keys, and then we calculate the general unlinkability measure introduced in \cite{gomez2017general}. 
The linkability of two templates is measured in terms of the difference of conditional probabilities of two hypotheses of being mated, $H_{m}$, and non-mated, $H_{nm}$, for a given comparison score $s$ between two given templates:
\begin{equation}
\mathrm{D}_{\leftrightarrow}(s)= p(H_m|s) - p(H_{nm}|s).
\end{equation}

Then, 
by finding conditional expectation of this local measure $\mathrm{D}_{\leftrightarrow}(s)$ over all comparison scores, 
we can find a global measure, $\mathrm{D}_{\leftrightarrow}^{\mathit{sys}}$, which is considered  as the system unlinkability metric:
\begin{equation}
    \mathrm{D}_{\leftrightarrow}^{\mathit{sys}} = {\int} p(s|H_m)  \mathrm{D}_{\leftrightarrow}(s) \text{d}s.
\end{equation}

The value of $\mathrm{D}_{\leftrightarrow}^{\mathit{sys}}$ is in interval [0,1], with lower values indicating smaller possibilities to link templates of the same subject.

\begin{figure*}[tb]
	\centering
	\hfill 
	\begin{subfigure}[b]{0.9325\textwidth}
		\centering
		\includegraphics[page=1,width=1.0\linewidth, trim={-0.03cm 0cm -.26cm 0cm},clip]{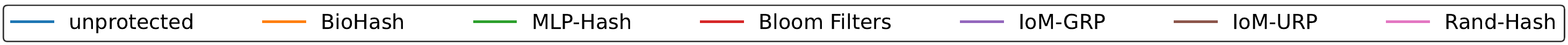}
	\end{subfigure}\hfill\\ 
	\rotatebox[]{90}{\small  \emph{normal} scenario \hspace{-110 pt}}\hspace{5pt}\hfil
	\begin{subfigure}[b]{0.24\textwidth}
		\centering
		\includegraphics[page=1,width=1.0\linewidth, trim={0cm 0cm 0cm 0cm},clip]{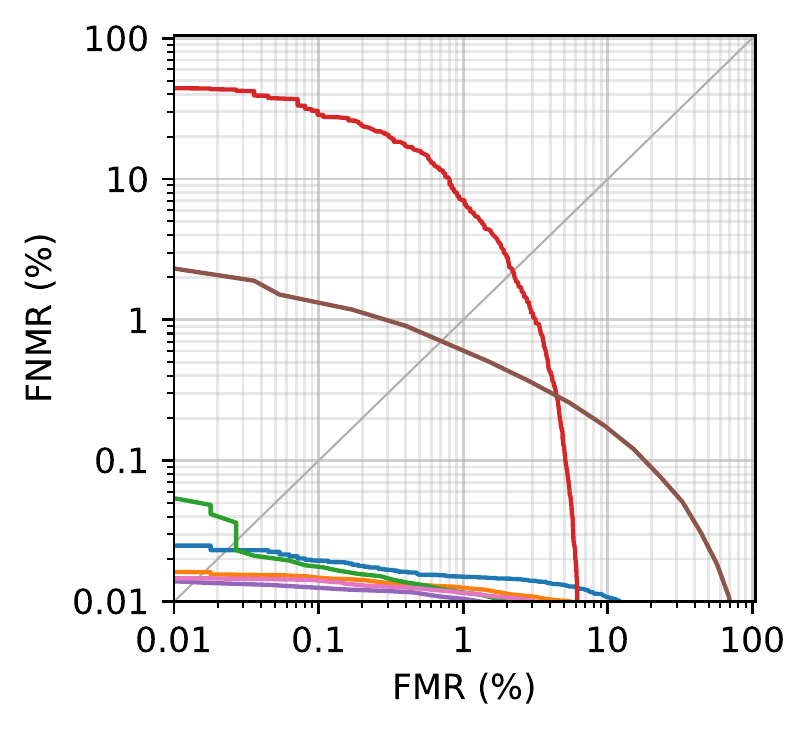}
	\end{subfigure}\hfil
	\begin{subfigure}[b]{0.24\textwidth}
		\centering
		\includegraphics[page=1,width=1.0\linewidth, trim={0cm 0cm 0cm 0cm},clip]{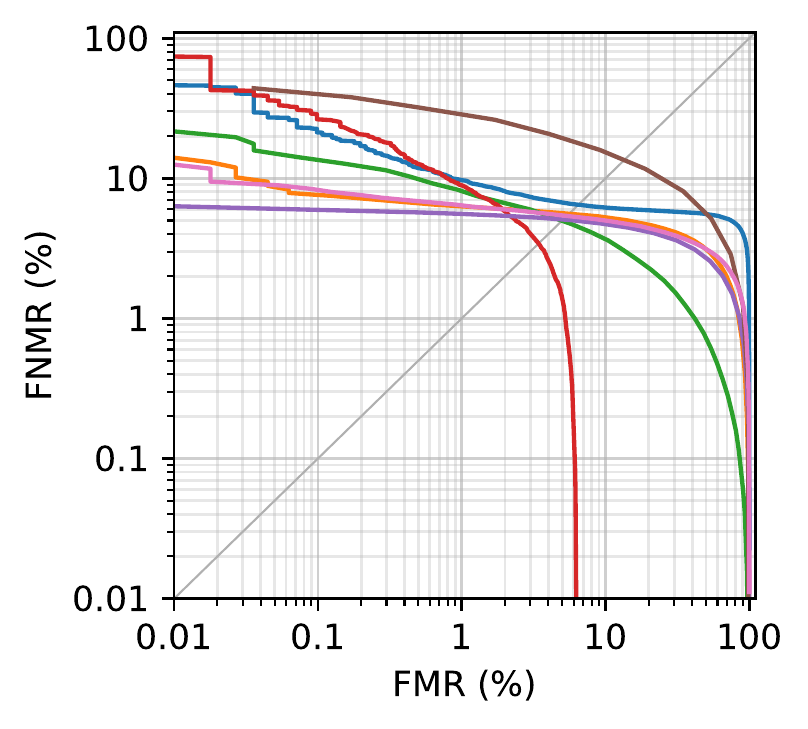}
	\end{subfigure}\hfil 
	\begin{subfigure}[b]{0.24\textwidth}
		\centering
		\includegraphics[page=1,width=1.0\linewidth, trim={0cm 0cm 0cm 0cm},clip]{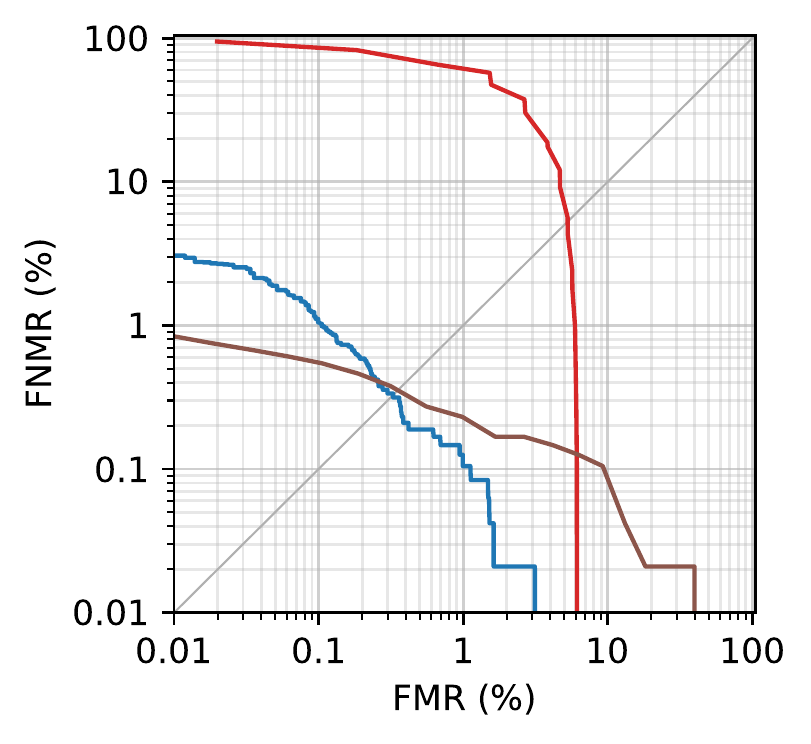}
	\end{subfigure}\hfil 
	\begin{subfigure}[b]{0.24\textwidth}
		\centering
		\includegraphics[page=1,width=1.0\linewidth, trim={0cm 0cm 0cm 0cm},clip]{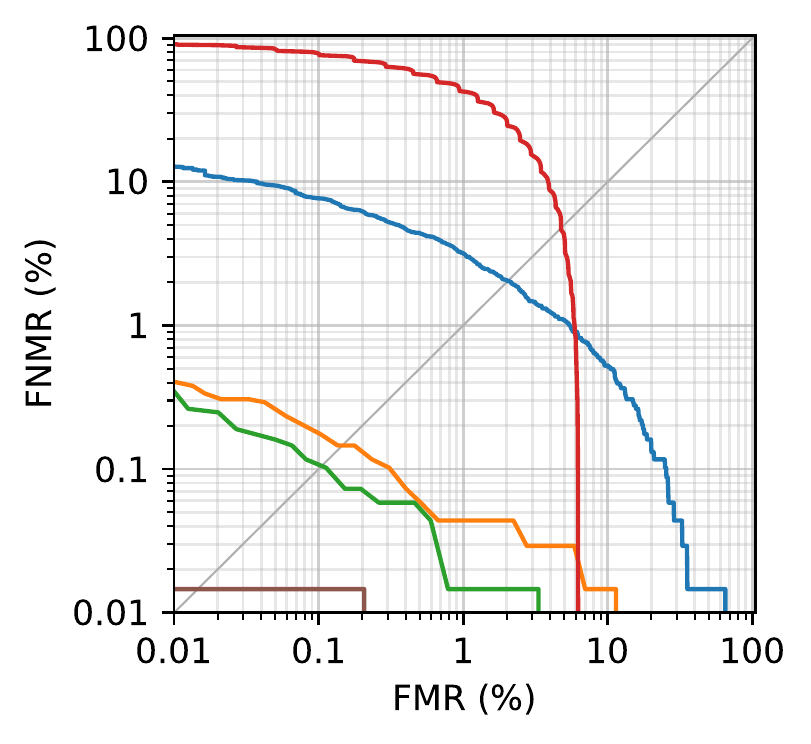}
	\end{subfigure}\hfil \\ \vspace{5pt}
	\rotatebox[]{90}{\small  \emph{stolen-token} scenario \hspace{-145 pt}}\hspace{5pt}\hfil
	\begin{subfigure}[b]{0.24\textwidth}
		\centering
		\includegraphics[page=1,width=1.0\linewidth, trim={0cm 0cm 0cm 0cm},clip]{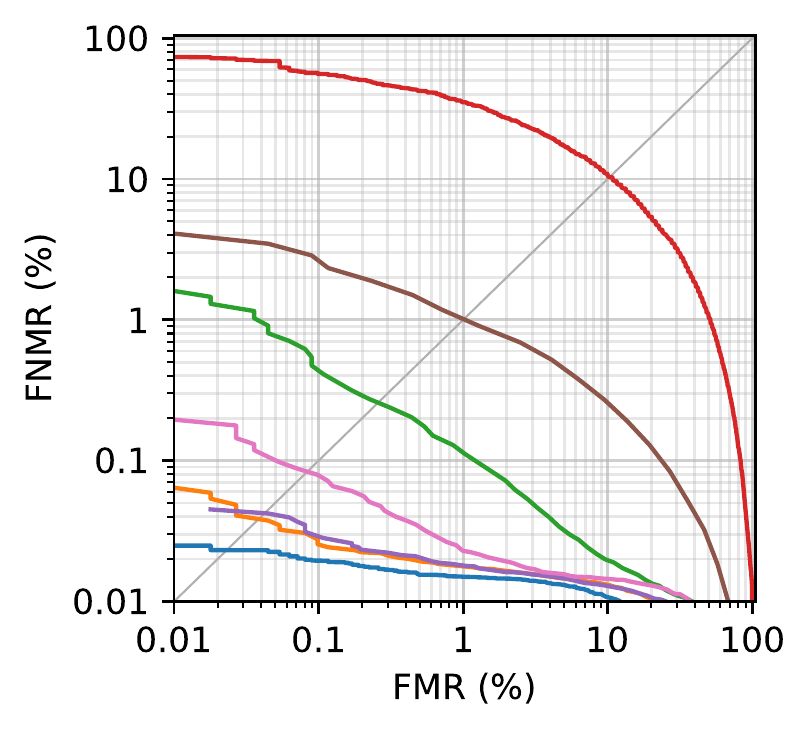}
		\caption{Face}
	\end{subfigure}\hfil
	\begin{subfigure}[b]{0.24\textwidth}
		\centering
		\includegraphics[page=1,width=1.0\linewidth, trim={0cm 0cm 0cm 0cm},clip]{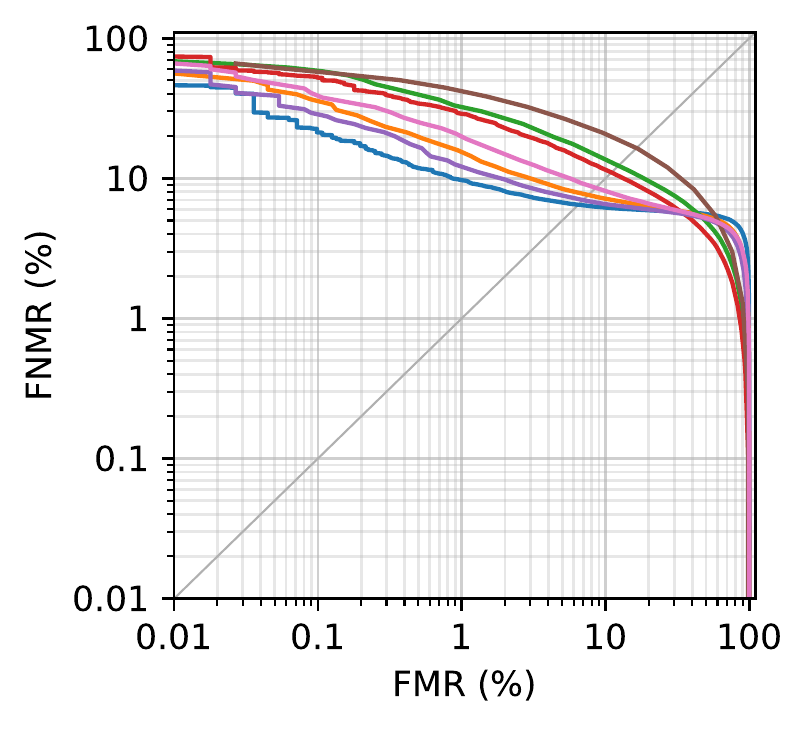}
		\caption{Voice}
	\end{subfigure}\hfil 
	\begin{subfigure}[b]{0.24\textwidth}
		\centering
		\includegraphics[page=1,width=1.0\linewidth, trim={0cm 0cm 0cm 0cm},clip]{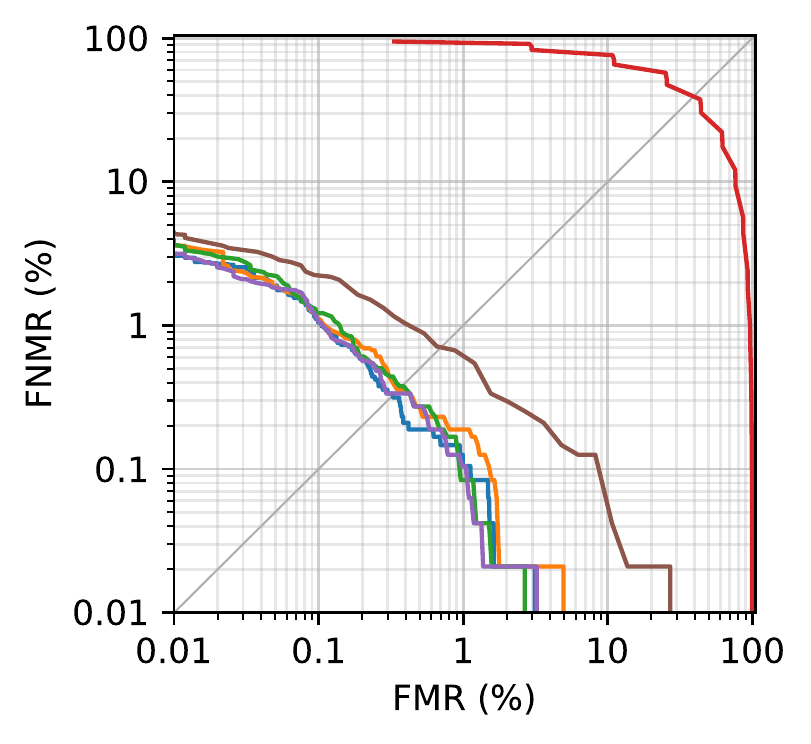}
		\caption{Finger Vein$^{*}$}
	\end{subfigure}\hfil 
	\begin{subfigure}[b]{0.24\textwidth}
		\centering
		\includegraphics[page=1,width=1.0\linewidth, trim={0cm 0cm 0cm 0cm},clip]{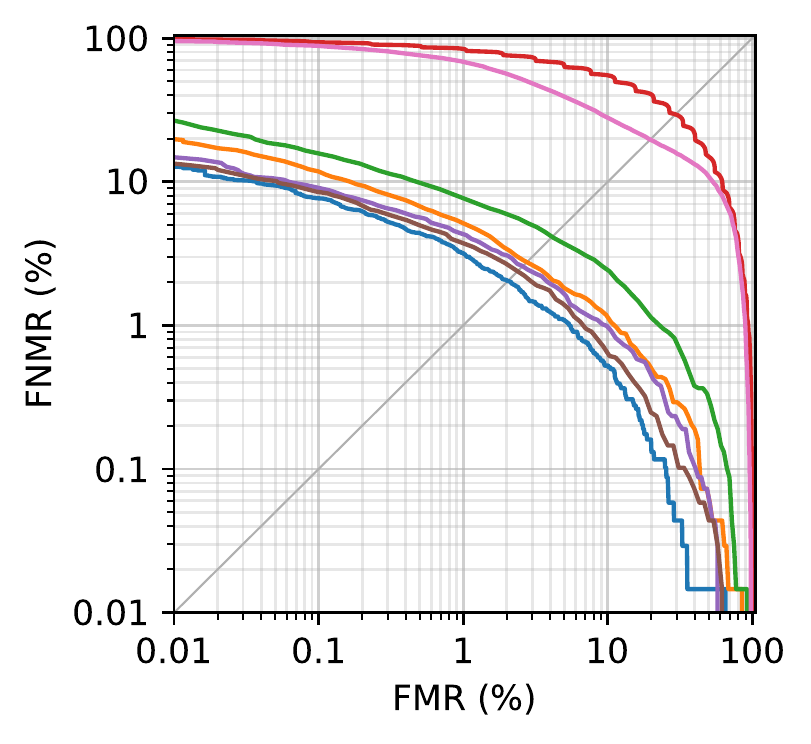}
		\caption{Iris$^{**}$}
	\end{subfigure}\hfil \\
	
	\flushleft\footnotesize{\qquad$^{*}$In the \textit{normal} scenario, DET curves of  protected templates with  BioHash, MLP-Hash, IoM-GRP, and Rand-Hash are not visible because EER=0. Similarly, in the \textit{stolen-token} scenario, DET curves of Rand-Hash is not visible.}\\\vspace{-3pt}
	\flushleft\footnotesize{\qquad$^{**}$In the \textit{normal} scenario, DET curve of protected templates with IoM-GRP is not visible because EER=0.}\\\vspace{-2pt}
	\caption{System performance evaluation on the \textit{normal} and \textit{stolen-token} scenarios for different physiological and behavioral biometric traits.}
	\label{fig:rec_performance}
	\vspace{-18pt}
\end{figure*}

\begin{table*}[tbp]
	\centering
	\setlength{\tabcolsep}{3pt}
	\caption{Recognition Performance Evaluation}
	\label{tab:performance}
	\scalebox{0.96}{
		\begin{tabular}[t]{@{} llcccccccc @{}}
			\toprule[1pt]
			\multirow{2}{*}{\textbf{CB scheme}} & \multirow{2}{*}{\textbf{Modality}}  & \multicolumn{3}{c}{\textbf{\textit{Normal} scenario [\%]}} &&   \multicolumn{3}{c}{\textbf{\textit{Stolen-token} scenario [\%]}}\\
			\cline{3-5}
			\cline{7-9}
			\rule{0pt}{2.5ex}    
			& & {EER} & {FNMR@FMR=1\%} & {FNMR@FMR=0.1\%} && {EER} & {FNMR@FMR=1\%} & {FNMR@FMR=0.1\%}  \\
			\midrule[1pt]
			\multirow{4}{*}{\textbf{Unprotected}} 
			&Face &  $0.02$ & $0.01$ & $0.02$  & & $-$ & $-$ &$-$    \\ \cline{2-9} \rule{0pt}{2.5ex}
			&Voice &  $6.4$ & $9.71$ & $22.53$  & & $-$ & $-$ &$-$    \\ \cline{2-9} \rule{0pt}{2.5ex}
			&Finger Vein &  $0.32$ & $0.10$ & $1.05$  & & $-$ & $-$ &$-$    \\ \cline{2-9} \rule{0pt}{2.5ex}
			&Iris&   $2.05$ & $3.18$ & $7.69$  & & $-$ & $-$ &$-$    \\ 
			\midrule
			\multirow{4}{*}{\textbf{BioHash} \cite{jin2004biohashing}} 
			&Face &  $0.02$ & $0.01$ & $0.01$  & & $0.04$ & $0.02$ & $0.03$    \\ \cline{2-9} \rule{0pt}{2.5ex}
			&Voice &  $5.28$ & $6.31$ & $8.31$  & & $7.64$ & $15.84$ & $36.60$    \\ \cline{2-9} \rule{0pt}{2.5ex}
			&Finger Vein &  $0.00$ & $0.00$ & $0.00$  & & $0.35$ & $0.19$ &$1.13$    \\ \cline{2-9} \rule{0pt}{2.5ex}
			&Iris&   $0.14$ & $0.04$ & $0.18$  & & $2.77$ & $5.15$ &$11.80$    \\ 
			\midrule
			\multirow{4}{*}{\textbf{MLP-Hash} \cite{MLPHash}} 
			&Face &  $0.02$ & $0.01$ & $0.02$  & & $0.02$ & $0.13$ & $0.54$    \\ \cline{2-9} \rule{0pt}{2.5ex}
			&Voice &  $5.25$ & $8.30$ & $14.19$  & & $12.16$ & $33.12$ & $61.70$    \\ \cline{2-9} \rule{0pt}{2.5ex}
			&Finger Vein &  $0.00$ & $0.00$ & $0.00$  & & $0.37$ & $0.08$ &$1.22$    \\ \cline{2-9} \rule{0pt}{2.5ex}
			&Iris&   $0.11$ & $0.01$ & $0.10$  & & $4.14$ & $7.91$ &$15.80$    \\ 
			\midrule
			\multirow{4}{*}{\textbf{Bloom Filters} \cite{rathgeb2013alignment}} 
			&Face &  $2.19$ & $7.10$ & $30.53$  & & $35.40$ & $56.67$ &$43.33$    \\ \cline{2-9} \rule{0pt}{2.5ex}
			&Voice &  $3.42$ & $9.01$ & $28.70$  & & $11.03$ & $28.97$ &$53.03$    \\ \cline{2-9} \rule{0pt}{2.5ex}
			&Finger Vein &  $4.97$ & $65.41$ & $91.28$  & & $36.49$ & $94.76$ &$94.76$    \\ \cline{2-9} \rule{0pt}{2.5ex}
			&Iris&   $4.69$ & $42.57$ & $78.15$  & & $29.26$ & $84.11$ &$93.17$    \\ 
			\midrule
			\multirow{4}{*}{\textbf{IoM-GRP} \cite{jin2017ranking}} 
			&Face &  $0.02$ & $0.01$ & $0.01$  & & $0.04$ & $0.02$ & $0.03$    \\ \cline{2-9} \rule{0pt}{2.5ex}
			&Voice &  $5.36$ & $5.58$ & $5.96$  & & $7.14$ & $12.57$ &$29.34$    \\ \cline{2-9} \rule{0pt}{2.5ex}
			&Finger Vein &  $0.00$ & $0.00$ & $0.00$  & & $0.33$ & $0.10$ &$1.05$    \\ \cline{2-9} \rule{0pt}{2.5ex}
			&Iris&   $0.00$ & $0.00$ & $0.00$  & & $2.58$ & $4.25$ &$9.25$    \\ 
			\midrule
			\multirow{4}{*}{\textbf{IoM-URP} \cite{jin2017ranking}} 
			&Face &  $0.72$ & $0.68$ & $1.51$  & & $1.10$ & $1.18$ & $2.87$  \\ \cline{2-9} \rule{0pt}{2.5ex}
			&Voice &  $12.35$ & $31.93$ & $43.91$  & & $16.52$ & $44.40$ &$61.14$    \\ \cline{2-9} \rule{0pt}{2.5ex}
			&Finger Vein &  $0.35$ & $0.23$ & $0.55$  & & $0.69$ & $0.67$ &$2.24$    \\ \cline{2-9} \rule{0pt}{2.5ex}
			&Iris&   $0.02$ & $0.00$ & $0.01$  & & $2.38$ & $3.66$ &$8.56$    \\
			\midrule
			\multirow{4}{*}{\textbf{Rand-Hash}} 
			&Face &  $0.02$ & $0.01$ & $0.01$  & & $0.08$ & $0.02$ & $0.08$   \\ \cline{2-9} \rule{0pt}{2.5ex}
			&Voice &  $5.70$ & $6.48$ & $8.40$ & & $8.35$ & $20.81$ & $40.66$    \\ \cline{2-9} \rule{0pt}{2.5ex}
			&Finger Vein &  $0.00$ & $0.00$ & $0.00$  & & $0.00$ & $0.00$ &$0.00$    \\ \cline{2-9} \rule{0pt}{2.5ex}
			&Iris&   $0.00$ & $0.00$ & $0.00$  & & $19.76$ & $68.11$ &$88.55$    \\
			\bottomrule[1pt]
			\vspace{-15pt}
		\end{tabular}\label{tab:rec_performance}
	}
	
\end{table*}

\begin{table}[]
\centering
\caption{Unlinkability Evaluation}
\begin{tabular}{lcccc}
\hline
\multicolumn{1}{c}{\textbf{CB scheme}} & \textbf{Face}        & \textbf{Voice}       & \textbf{Finger Vein} & \textbf{Iris}        \\ \hline
\textbf{BioHash} \cite{jin2004biohashing}          & $0.0110$ &   $0.0078$  & $0.0140$                      & $0.0106$\\ \hline
\textbf{MLP-Hash} \cite{MLPHash}             & $0.0088$  &     $0.0160$       &$0.0099$                      & $0.0122$                     \\ \hline
\textbf{Bloom Filter} \cite{rathgeb2013alignment}         & $0.0545$ & $0.0735$  &$0.0091$        &  $0.0131$\\ \hline
\textbf{IoM-GRP} \cite{jin2017ranking}                & $0.0086$ &     $0.0065$   &$0.0130$                      &   $0.0072$                   \\ \hline
\textbf{IoM-URP} \cite{jin2017ranking}                & $0.0090$ &     $0.0053$    &$0.0136$                      &   $0.0090$                   \\ \hline
\textbf{Rand-Hash}           & $0.0084$ &      $0.0061$      &$0.0107$                      &     $0.0099$                 \\ \hline
\vspace{-17pt}
\end{tabular}\label{tab:unlinkability}
\end{table}

\begin{table}[tbp]
	\centering
	\setlength{\tabcolsep}{3pt}
	\caption{Irreversiblity Evaluation}
	\label{tab:irreversibility}
	\scalebox{.99}{
		\begin{tabular}[t]{@{} llcc @{}}
		\toprule[1pt]
			\textbf{CB scheme} & \textbf{Characteristic}  & \textbf{\begin{tabular}[c]{@{}c@{}}\emph{Normal}\\ scenario\end{tabular}} &  \textbf{\begin{tabular}[c]{@{}c@{}}\emph{Stolen-token}\\ scenario\end{tabular}} \\ 
			\midrule[1pt]
			\multirow{4}{*}{\textbf{BioHash} \cite{jin2004biohashing}} 
			&Face & $39.63$  & $98.81$    \\ \cline{2-4} \rule{0pt}{2.5ex}
			&Voice & $12.97$  & $53.74$    \\ \cline{2-4} \rule{0pt}{2.5ex}
			&Finger Vein & $18.80$  & $115.99$    \\ \cline{2-4} \rule{0pt}{2.5ex}
			&Iris& $8.63$  & $63.99$    \\ 
			\midrule
			\multirow{4}{*}{\textbf{MLP-Hash} \cite{MLPHash}} 
			&Face & $35.42$  & $58.00$    \\ \cline{2-4} \rule{0pt}{2.5ex}
			&Voice & $10.74$  &$26.37$    \\ \cline{2-4} \rule{0pt}{2.5ex}
			&Finger Vein & $19.35$  &$110.04$    \\ \cline{2-4} \rule{0pt}{2.5ex}
			&Iris & $7.92$  &$38.65$   \\ 
			\midrule
			\multirow{4}{*}{\textbf{Bloom Filters} \cite{rathgeb2013alignment}} 
			&Face & $40.18$  & $21.37$    \\ \cline{2-4} \rule{0pt}{2.5ex}
			&Voice & $20.26$  & $29.60$    \\ \cline{2-4} \rule{0pt}{2.5ex}
			&Finger Vein & $12.32$  & $8.89$    \\ \cline{2-4} \rule{0pt}{2.5ex}
			&Iris & $8.14$  &$8.56$   \\ 
			\midrule
			\multirow{4}{*}{\textbf{IoM-GRP} \cite{jin2017ranking}} 
			&Face & $31.33$  & $48.91$    \\ \cline{2-4} \rule{0pt}{2.5ex}
			&Voice & $8.68$  & $22.83$    \\ \cline{2-4} \rule{0pt}{2.5ex}
			&Finger Vein & $18.18$  & $57.85$    \\ \cline{2-4} \rule{0pt}{2.5ex}
			&Iris & $8.31$  & $38.29$    \\ 
			\midrule
			\multirow{4}{*}{\textbf{IoM-URP} \cite{jin2017ranking}} 
			&Face & $8.79$  & $9.06$    \\ \cline{2-4} \rule{0pt}{2.5ex}
			&Voice & $1.69$ &$3.10$    \\ \cline{2-4} \rule{0pt}{2.5ex}
			&Finger Vein & $14.01$  & $16.06$    \\ \cline{2-4} \rule{0pt}{2.5ex}
			&Iris & $6.55$ &$24.59$    \\
			\midrule
			\multirow{4}{*}{\textbf{Rand-Hash}} 
			&Face & $39.26$  & $97.07$    \\ \cline{2-4} \rule{0pt}{2.5ex}
			&Voice & $12.47$  & $52.31$    \\ \cline{2-4} \rule{0pt}{2.5ex}
			&Finger Vein & $19.65$  & $113.96$    \\ \cline{2-4} \rule{0pt}{2.5ex}
			&Iris& $9.77$  & $26.43$    \\
			\bottomrule[1pt]
		\end{tabular}
	}\vspace{-12pt}
	
\end{table}

\subsubsection{Irreversibility}
Although many information-theoretic metrics have been proposed in the literature, a general framework for the evaluation of irreversibility is currently missing \cite{melzi2022}. One of the most popular metrics for irreversibility is mutual information (MI). It quantifies the amount of information related to the set of original (unprotected) biometric templates $X$ that can be obtained from the set of protected biometric templates $Y$. The set $Y$ is obtained with the application of some CB schemes to the set of unprotected templates $X$. We consider $Y$ available to attackers, under both the \emph{normal} and \emph{stolen} scenarios. The calculation of MI requires in input the two sets of unprotected and protected templates, and provides a non-negative score in output. The smaller this score, the better for irreversibility, with value equal to zero when the two sets are independent, i.e. no information about the original templates can be disclosed by the protected templates. The computation of MI relies on the estimation of entropy, which is hard to compute, especially if biometric templates contain an high number of features \cite{nandakumar2015biometric}. 

To simplify the computation of entropy and MI, we apply Principal Component Analysis (PCA) to our sets $X$ and $Y$, that are matrices with initial dimensions $s \times u$ and $s \times p$ respectively, with $s$ being the number of samples, $u$ the number of features in unprotected templates, and $p$ the number of features in protected templates. From the application of PCA to the matrices $X$ and $Y$, we obtain the reduced matrices $X_r = PCA(X)$ and $Y_r = PCA(Y)$, with dimensions $s \times r$, where $r$ is the number of reduced features (possibly different across matrices). While decreasing the number of features, PCA retains the most significant information of biometric templates. That is, reduced matrices are suitable to account for the partial reversibility of protected biometric data, which in many cases is sufficient to obtain access in biometric recognition systems. 

To obtain a fair comparison between the different CB schemes, we apply PCA to the matrices of unprotected templates $X$ and protected templates $Y_i$ resulting from different CB schemes $i$, always considering a fixed number of features $r=100$ for the reduced matrices. Then, we approximate to multivariate Gaussian the distribution of features of the reduced matrices. For each matrix ${Y_r}_i$, we can compute the MI between $X_r$ and ${Y_r}_i$ as follows:
\begin{equation}
MI(X_r, {Y_r}_i) = H(X_r) + H({Y_r}_i) - H(X_r, {Y_r}_i),
\end{equation}
where $H(\cdot)$ is the measure of entropy \cite{shannon1948mathematical}.

\section{Experiments}\label{sec:experiments}
In this section, we report the experimental results of our benchmark framework for the evaluation of different CB schemes: BioHashing \cite{jin2004biohashing}, Multi-Layer Perceptron (MLP) Hashing \cite{MLPHash}, Bloom Filters \cite{rathgeb2013alignment}, and two schemes based on Index-of-Maximum (IoM) Hashing \cite{jin2017ranking}   (i.e., IoM-URP, and IoM-GRP).
In addition, as a baseline, we consider a CB scheme named Rand-Hash, based on user-specific random transformations (including random permutation, random scale, and random sign flip) followed by binarization.

For our experiments on face and voice data, we use the Bob\footnote{Available at  \url{https://www.idiap.ch/software/bob/}} toolbox \cite{bob2012,bob2017}. To implement  the  BioHashing, MLP-Hashing, IoM-GRP, IoM-URP, we use the open-source  implementation of these BTP schemes in Bob \cite{hatef_TBIOM2021,shahreza2021deep,shahreza2021recognition,MLPHash}. 
The source code from our experiments is publicly available to facilitate the  reproducibility of our results\footnote{Source code: \url{https://github.com/otroshi/benchmark_cb}}.

\vspace{-3pt}
\subsection{Recognition Performance Evaluation}
Figure \ref{fig:rec_performance} depicts the DET curves for different CB schemes on different biometric characteristics. Table also \ref{tab:rec_performance}  compares the recognition performance in terms of Equal Error Rate (EER) as well as False Non-Match Rate (FNMR) at a False Match Rate (FMR) of $1\%$ and $0.1\%$. In general, Bloom Filters (which was not initially proposed to protect DNN-based features) has the lowest recognition performance for face, finger vein, and iris. For voice recognition however, IoM-URP has the worst recognition performance. Also, we observe that in face recognition, BioHash, MLP-Hash, IoM-GRP, IoM-URP, and Rand-Hash have comparable performance in the \textit{normal} scenario. In voice recognition, IoM-GRP achieves the best performance in the \textit{normal} and \textit{stolen} scenarios. In finger vein recognition, BioHash, MLP-Hash, IoM-GRP, IoM-URP, and Rand-Hash have comparable performance in the \textit{normal} scenario. Nevertheless, Rand-Hash achieves the best recognition performance in the \textit{stolen} scenario for finger vein recognition.  
Similarly in iris recognition, we observe that BioHash, MLP-Hash, IoM-GRP, IoM-URP, and Rand-Hash have comparable performance, and IoM-GRP achieves the best performance. However, in the \textit{stolen} scenario, IoM-URP achieves the best recognition performance for iris recognition. It is noteworthy that generally for each biometric characteristic, protected templates with some of the CB schemes scenario achieve better recognition accuracy than unprotected templates in the \textit{normal}. The improvement in the accuracy in such cases is obtained with the cost of using  user-specific keys.

\vspace{-2pt}
\subsection{Unlinkability Evaluation}\vspace{-3pt}
Table \ref{tab:unlinkability} compares the unlinkability metric for  different CB schemes when protecting different biometric modalities. 
This metric evaluates unlinkability of protected templates based on the overlap between the distribution of scores of  mated templates and the distribution of scores of non-mated templates protected with different keys. Therefore,
if the distribution of scores of mated templates and the distribution of scores  non-mated templates largely overlap, 
the global measure $\mathrm{D}_{\leftrightarrow}^{\mathit{sys}}$ will be close to zero. Therefore, based on the hypothesis test in this measure, it is unfeasible to link templates and, hence, protected templates are considered to be unlinkable.
Accordingly, as Table \ref{tab:unlinkability} shows, all CB schemes are almost unlinkable across different biometric characteristics. 

\vspace{-2pt}
\subsection{Irreversibility Evaluation}\vspace{-3pt}
In Table \ref{tab:irreversibility}, for each biometric characteristic, CB scheme, and scenario, we report the MI between the reduced matrices of unprotected and protected templates. By comparing the MI values obtained between the \textit{normal} and \textit{stolen} scenarios, we observe a clear increase of MI in the \textit{stolen} scenario. In the latter scenario, the key required by CB schemes is no more user-specific and it can be simply considered as a parameter of the CB scheme. As a consequence, from protected templates in the \textit{stolen} scenario it is easier to extract information about the original (unprotected) templates. We also observe that, in general, for the \textit{normal} scenario face is the biometric characteristic that provides the highest MI, while for the \textit{stolen} scenario finger vein is the biometric characteristic that provides the highest MI. 

\section{Conclusion}\label{sec:conclusion}
In this paper, we presented a comprehensive benchmark by evaluating the recognition performance, unlinkability, and irreversibility of deep templates from different biometric characteristics, which are protected with different CB schemes. We used SOTA DNN models to extract features from face, voice, finger vein, and iris, and evaluated their protected templates using BioHashing, MLP-Hashing, Bloom Filters, IoM-URP, and IoM-GRP. In addition to the mentioned CB schemes, we introduced a CB scheme named Rand-Hash based on user-specific random transformations followed by binarization.
Our experiments show that all the mentioned CB schemes achieve close to perfect unlinkability across different characteristics. 
We also evaluate the irreversibility in terms of MI. We observe that the computed MI varies according to the scenario and biometric characteristic considered. In particular, in the \textit{stolen} scenario the MI between unprotected and protected templates is higher than the corresponding values in \textit{normal} scenario.
Last but not least, it was found that when applied to deep templates Bloom Filter-based protection causes a drop in recognition performance, but other CB schemes achieve competitive performance across different characteristics.

\section*{Acknowledgment}\label{sec:Acknowledgment}
This research is based upon work supported by the H2020 TReSPAsS-ETN Marie Sk\l{}odowska-Curie early training network (grant agreement 860813) and by the German Federal Ministry of Education and Research and the Hessian Ministry of Higher Education, Research, Science and the Arts within their joint support of the National Research Center for Applied Cybersecurity ATHENE.
\vspace{-3pt}
\bibliographystyle{IEEEtran}
\bibliography{refs}

\end{document}